\documentclass[acmtog,authorversion]{acmart}
\acmSubmissionID{392}

\usepackage{booktabs} 

\setcitestyle{square}

\usepackage[ruled]{algorithm2e} 

\SetAlFnt{\small}
\SetAlCapFnt{\small}
\SetAlCapNameFnt{\small}
\SetAlCapHSkip{0pt}
\IncMargin{-\parindent}

\setcopyright{none}

\acmDOI{0000001.0000001_2}

\usepackage{color}
\usepackage{soul}
\usepackage[ruled]{algorithm2e} 
\usepackage{rotating}

\definecolor{purple}{rgb}{0.65,0,0.65}
\definecolor{blue}{rgb}{0, 0.2, 0.8}
\definecolor{orange}{rgb}{0.6, 0.6, 0}
\definecolor{red}{rgb}{0.8, 0.2, 0.2}
\definecolor{magenta}{rgb}{0.5, 0.0, 1.0}
\definecolor{black}{rgb}{0.0, 0.0, 0.0}
\definecolor{cyan}{rgb}{0, 0.65, 0.65}
\usepackage{xcolor}

\newif\ifdraft
\draftfalse

\ifdraft
\newcommand{\dlc}[1]{{\color{blue}\textbf{DL:} #1}}
\newcommand{\dcc}[1]{{\color{red}\textbf{DC:} #1}}
\newcommand{\kac}[1]{{\color{orange}\textbf{KA:} #1}}
\newcommand{\nfc}[1]{{\color{green}\textbf{NF:} #1}}

\else
\newcommand{\dlc}[1]{}
\newcommand{\dcc}[1]{}
\newcommand{\kac}[1]{}
\newcommand{\nfc}[1]{}

\fi
\newcommand{\bbd}{{\bf D}}

\newcommand{\bbg}{{\bf G}}

\newcommand{\bbp}{{\bf P}}
\newcommand{\bbw}{{\bf W}}
\newcommand{\Loss}{\mathcal{L}}
\newcommand{\Dp}{\bbd_{\text{p}}}
\newcommand{\Ds}{\bbd_{\text{s}}}
\newcommand{\pp}{\mathcal{P}}
\newcommand{\bbe}{\mathbb{E}}
\def \figures {./}

\begin{document}
\title{Deep Video-Based Performance Cloning}

\author{Kfir Aberman}
\affiliation{%
\institution{AICFVE Beijing Film Academy}
}

\author{Mingyi Shi}
\affiliation{%
  \institution{Shandong University}
}  

\author{Jing Liao}
\affiliation{%
  \institution{Microsoft Research Asia}
}

\author{Dani Lischinski}
\affiliation{%
  \institution{Hebrew University of Jerusalem}
  }
  \author{Baoquan Chen}
\affiliation{%
  \institution{AICFVE Beijing Film Academy,}
  \institution{Peking University}
  }
  \author{Daniel Cohen-Or}
\affiliation{%
  \institution{Tel-Aviv University}
}

\renewcommand\shortauthors{Aberman, K. et al}

\begin{abstract}
We present a new video-based performance cloning technique. After training a deep generative network using a reference video capturing the appearance and dynamics of a target actor, we are able to generate videos where this actor reenacts other performances. All of the training data and the driving performances are provided as ordinary video segments, without motion capture or depth information.
Our generative model is realized as a deep neural network with two branches, both of which train the same space-time conditional generator, using shared weights. One branch, responsible for learning to generate the appearance of the target actor in various poses, uses \emph{paired} training data, self-generated from the reference video. The second branch uses \emph{unpaired} data to improve generation of temporally coherent video renditions of unseen pose sequences.
We demonstrate a variety of promising results, where our method is able to generate temporally coherent videos, for challenging scenarios where the reference and driving videos consist of very different dance performances.
\\Supplementary video: \url{https://youtu.be/JpwsEeqNhhA}
\end{abstract}

%
%

%
%



\begin{teaserfigure}
\centering
\includegraphics[width=\linewidth]{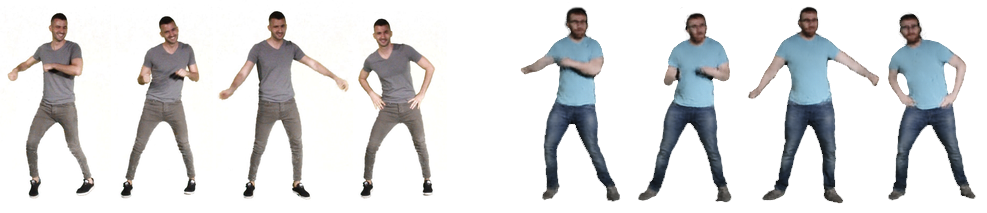} 
\caption{Video-driven performance cloning transfers the motions of a performer in a driving video (left) to a target actor (right), captured in a reference video.}
\label{fig:teaser}
\end{teaserfigure}
 
%




\maketitle

\section{Introduction}
\label{sec:intro}
Wouldn't it be great to be able to have a video of yourself moonwalking like Michael Jackson, or moving like Mick Jagger? Unfortunately, not many of us are capable of delivering such exciting and well-executed performances. However, perhaps it is possible to use an existing video of Jackson or Jagger to convert a video of an imperfect, or even a completely unrelated, reference performance into one that looks more like ``the real deal''? In this paper, we attempt to provide a solution for such a scenario, which we refer to as \emph{video-driven performance cloning}.

Motion capture is widely used to drive animation. However, in this scenario, the target to be animated is typically an articulated virtual character, which is explicitly rigged for animation, rather than a target individual captured by a simple RGB video. Video-driven performance cloning, which transfers the motion of a character between two videos, has also been studied.
Research has mainly focused on facial performance reenactment, where speech, facial expressions, and sometimes changes in head pose, are transferred from one video to another (or a to a still image), while keeping the identity of the target face. To accomplish the transfer, existing works typically leverage dense motion fields and/or elaborate controllable parametric facial models, which are not yet available for the full human body. Cloning full body motion remains an open challenging problem, since the space of clothed and articulated humans is much richer and more varied than that of faces and facial expressions.

The emergence of neural networks and their startling performance and excellent generalization capabilities have recently driven researchers to develop generative models for video and motion synthesis. In this paper, we present a novel two-branch generative network that is trained to clone the full human body performance captured in a \emph{driving video} to a target individual captured in a given \emph{reference video}.

Given an ordinary video of a performance by the target actor, its frames are analyzed to extract the actor's 2D body poses, each of which is represented as a set of part confidence maps, corresponding to the joints of the body skeleton~\cite{cao2017realtime}.
The resulting pairs of video frames and the extracted poses are used to train a generative model to translate such poses to frames, in a self-supervised manner.
Obviously, the reference video contains only a subset of the possible poses and motions that one might like to generate when cloning another performance. Thus, the key idea of our approach is to improve the generalization capability of our model by also training it with unpaired data.
The unpaired data consists of pose sequences extracted from other relevant videos, which could depict completely different individuals.

The approach outlined above is realized as a deep network with two branches, both of which train the same space-time conditional generator (using shared weights). One branch uses \emph{paired} training data with a combination of reconstruction and adversarial losses. The second branch uses \emph{unpaired} data with a combination of temporal consistency loss and an adversarial loss.
The adversarial losses used in this work are patchwise (Markovian) making it possible to generate new poses of the target individual by ``stitching together'' pieces of poses present in the reference video.

Following a review of related work in Section~\ref{sec:related}, we present are poses-to-video translation network in Section~\ref{sec:network}. Naturally, the success of our approach depends greatly on the richness of the paired training data (the reference video), the relevance of the unpaired pose sequences used to train the unpaired branch, and the degree of similarity between poses in the driving video and those which were used to train the network. In Section~\ref{sec:metric} we define a similarity metric between poses, which is suitable for our approach, and is able to predict how well a given driving pose might be translated into a frame depicting the target actor.

Section~\ref{sec:results} describes and reports some qualitative and quantitative results and comparisons. Several performance cloning examples are included in the supplementary video. The cloned performances are temporally coherent, and preserve the target actor's identity well. Our results exhibit some of the artifacts typical of GAN-generated imagery. Nevertheless, we believe that
they are quite promising, especially considering some of the challenging combinations of reference and driving videos that they have been tested with.

\section{Related Work}
\label{sec:related}

\subsubsection*{\textbf{Image Translation}}

Conditional GANs \cite{mirza2014conditional} have proven effective for image-to-image translation tasks which are aimed at converting between images from two different domains.
The structure of the main generator often incorporates an encoder-decoder architecture \cite{hinton2006reducing}, with skip connections \cite{ronneberger2015u}, and adversarial loss functions \cite{goodfellow2014generative}.
Isola et al.~\shortcite{isola2017image} demonstrate impressive image translation results obtained by such an architecture, trained using paired data sets.
Corresponding image pairs between different domains are often not available, but in some cases it is possible to use a translation operator that given an image in one domain can automatically produce a corresponding image from another.
Such self-supervision have been successfully used for tasks such as super-resolution \cite{Ledig2016}, colorization \cite{Zhang2016}, etc.
Recent works show that it is even possible to train high-resolution GANs \cite{karras2017progressive} as well as conditional GANs \cite{wang2017high}. 

However, in many tasks, the requirement for paired training data poses a significant obstacle. This issue is tackled by CycleGAN \cite{zhu2017unpaired}, DualGAN \cite{yi2017dualgan}, UNIT \cite{liu2017unsupervised}, and Hoshen and Wolf~\shortcite{hoshen2018identifying}. These unsupervised image-to-image translation techniques only require two sets of unpaired training samples. 

In this work, we are concerned with translation of 2D human pose sequences into photorealistic videos depicting a particular target actor.
In order to achieve this goal we propose a novel combination of self-supervised paired training with unpaired training, both of which make use of (different) customized loss functions.

\subsubsection*{\textbf{Human Pose Image Generation}}
Recently, a growing number of works aim to synthesize novel poses or views of humans, while preserving their identity.
Zhao et al.~\shortcite{zhao2017multi} learn to generate multiple views of a clothed human from only a single view input by combining variational inference and GANs.
This approach is conditioned by the input image and the target view, and does not provide control over the articulation of the person. 

In contrast, Ma et al.~\shortcite{ma2017pose} condition their person image generation model on both a reference image and the specified target pose. Their two-stage architecture, trained on pairs of images capturing the same person in different poses, first generates a coarse image with the target pose, which is then refined.
Siarohin et al.~\shortcite{siarohin2017deformable} attempt to improve this approach and extend it to other kinds of deformable objects for which sufficiently many keypoints that capture the pose may be extracted.

The above works require paired data, do not address temporal coherence, and in some cases their ability to preserve identity is limited, because of reliance on a single conditional image of the target individual.
In contrast, our goal is to generate a temporally coherent photorealistic video that consistently preserves the identity of a target actor. Our approach combines self-supervised training using paired data extracted from a video of the target individual with unpaired data training, and explicitly accounts for temporal coherence. 

Ma et al.~\shortcite{ma2017disentangled} propose to learn a representation that disentangles background, foreground, and pose. In principle, this enables modifying each of these factors separately. However, new latent features are generated by sampling from Gaussian noise, which is not suitable for our purpose and does not consider temporal coherence.

\subsubsection*{\textbf{Deep Video Generation}}
Video generation is a highly challenging problem, which has recently been tackled with deep generative networks. 
Vondrick et al.~\shortcite{vondrick2016generating} propose a generative model that predicts plausible futures from static images using a GAN consisting of 3D deconvolutional layers, while Saito et al.~\shortcite{saito2017temporal} use a temporal GAN that decomposes the video generation process into a temporal generator and an image generator.
These methods do not support detailed control of the generated video, as required in our scenario.

Walker et al.~\shortcite{walker2017pose} attempt to first predict the motion of the humans in an image, and use the inferred future poses as conditional information for a GAN that generates the frame pixels. In contrast, we are not concerned with future pose prediction; the poses are obtained from a driving video, and we focus instead on the generation of photorealistic frames of the target actor performing these poses.

The MoCoGAN \cite{tulyakov2017mocogan} video generation framework maps random vectors to video frames, with each vector consisting of a content part and a motion part. This allows one to generate videos with the same motion, but different content. In this approach the latent content subspace is modeled using a Gaussian distribution, which does not apply to our scenario, where the goal is to control the motion of a specific target individual.

\subsubsection*{\textbf{Performance Reenactment}}
Over the years, multiple works addressed video-based facial reenactment, where pose, speech, and/or facial expressions are transferred from a source performance to an image or a video of a target actor, e.g., \cite{averbuch2017,Suwajanakorn2015,Thies2016face2face,Vlasic2005,Dale2011,Kemelmacher2010,Garrido2014}.
These works typically rely on dense motion fields or leverage the availability of detailed and fully controllable  elaborate parametric facial models. Such models are not yet available for full human bodies, and we are not aware of methods capable of performing full body motion reenactment from video data.


Methods for performance driven animation \cite{xia2017survey}, use a video or a motion-captured performance to animate 3D characters. In this scenario, the target is typically an articulated 3D character, which is already rigged for animation. In contrast, our target is a real person captured using plain video. 

\section{Poses-to-Video Translation Network}
\label{sec:network}
The essence of our approach is to learn how to translate coherent sequences of 2D human poses into video frames, depicting an individual whose appearance closely resembles a \emph{target actor} captured in a provided \emph{reference video}.
Such a translator might be used, in principle, to clone arbitrary video performances of other individuals by extracting their human poses from various \emph{driving videos} and feeding them into the translator.

\begin{figure*}
	\centering
	\includegraphics[width=0.8\linewidth]{\figures 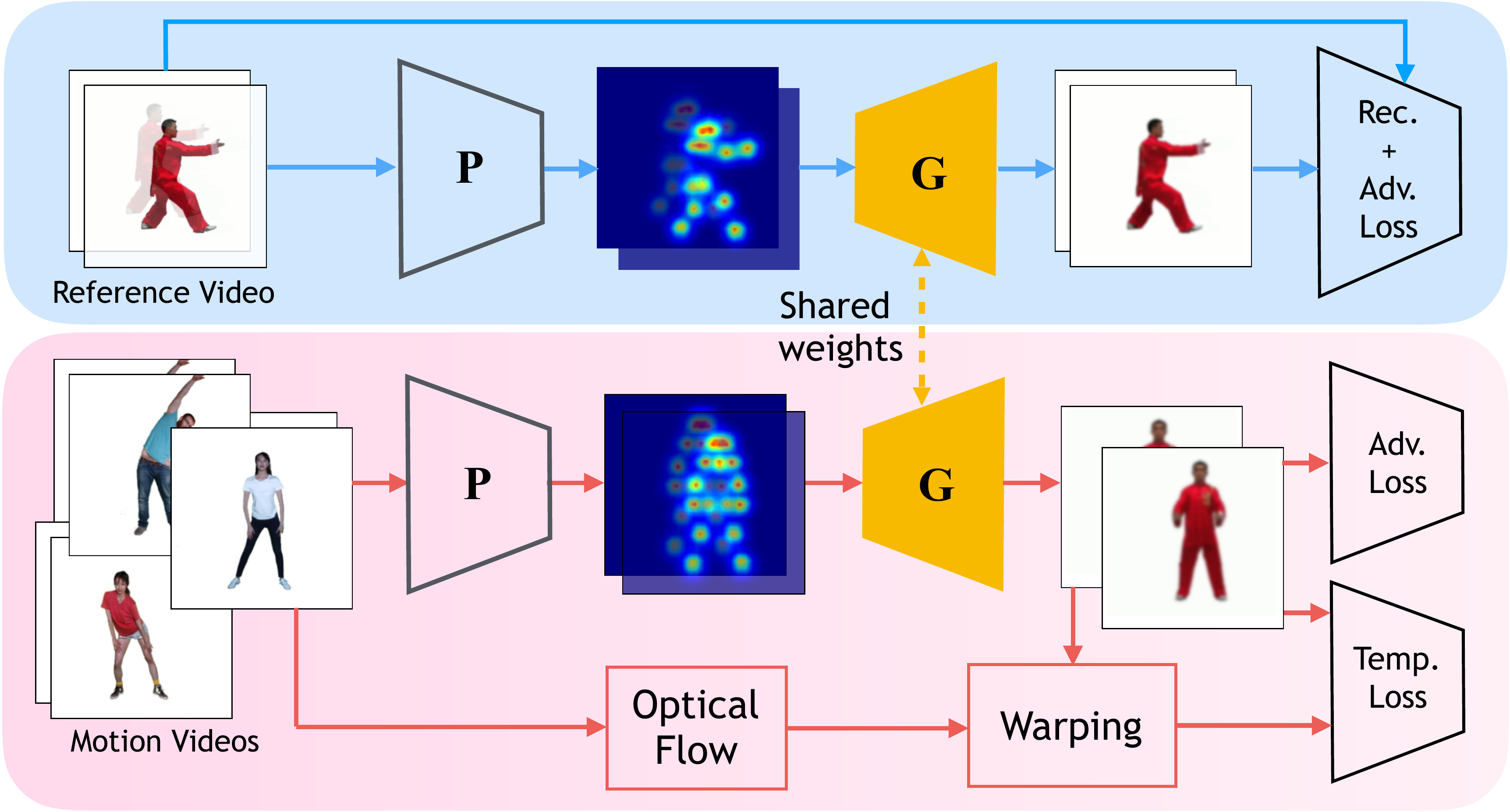}
	\caption{Our architecture trains a single generative network $\bbg$ using two training branches. The \emph{paired} branch (blue) is aimed at learning to map poses to frames, using a reconstruction loss on paired training data extracted from a reference video. The \emph{unpaired} branch (red) learns to translate unpaired poses into temporal coherent frames depict the target performer executing these poses. This is done using adversarial loss that takes care of maintaining the identity of the target actor and a temporal coherence loss requiring that consecutive frames will be temporal coherent, based on the optical flow of the corresponding driving video frames.}
	\label{fig:arc}
\end{figure*}

More specifically, we are given a reference video, capturing a range of motions of a target actor, whom we'd like to use in order to clone a performance of another individual, as well as one or more videos of other individuals.
We first train a deep neural network using \emph{paired data}: pairs of consecutive frames and their corresponding poses extracted from the reference video.
Since any given reference video contains only a subset of the poses and the motions that one might like to be able to generate, we gradually improve the generalization capability of this network by training it with \emph{unpaired data}, consisting of pairs of consecutive poses extracted from videos of other individuals, while requiring temporal coherency between consecutive frames.
An overview of our architecture is shown in Figure \ref{fig:arc}.

\subsection{Paired and Unpaired Data}
\label{subsec:data}
Let $X$ and $Y$ denote the domains of video frames and human poses, respectively, and let $\bbp:X\rightarrow Y$ be an operator which extracts the pose of a human in a given image. We first generate the paired data set that consists of frames  of a target actor, $x_i\in X$, and their corresponding estimated poses, $\bbp(x_i)\in Y$.  In addition, we are given some additional sequences of poses, $y_i\in Y$, which were extracted from additional videos. While the frames from which these poses were extracted are available as well, they typically capture other individuals, and frames of the target individual are not available for these poses.

The goal of our network, described in more detail in the next section, is to learn the mapping between arbitrary pose sequences to videos that appear to depict the target performer. The mapping is first learned using the paired data, and then extended to be able to translate the unpaired poses as well, in a temporally coherent manner.

\paragraph{Pose Representation}
To realize $\bbp$ we use the 2D human pose extraction method presented by Cao et al.~\shortcite{cao2017realtime}. In this method, 2D poses are represented by 3D volumes in $\mathbb{R}^{H\times W\times J}$, where $H$, $W$ are the height and width of the input image and $J$ is the number of channels, one per joint. Every channel is a part confidence map, containing a 2D Gaussian centered at the estimated location of a specific joint. There are 18 joints that constitute a basic skeleton. In addition to standard joints, such as elbows, knees, and shoulders, there are also joints for ears and eyes, so the head pose can be also roughly estimated. The part confidence maps explicitly provide the network with a separate spatial representation of each joint, in contrast to a flattened representation, such as a single map of the entire skeleton.
It should be noted that we roughly align the videos so that the average center point of the character (in time) will be positioned at the center of the frames and scale their size by fixing their average hips size to the same length (one scale and one translation for every video). In addition, we normalize each joint channel to have zero mean and unit variance, in order to achieve similar map values for different characters. 

%
%

\subsection{Network Architecture}
The core of our network consists of two branches, both of which train the same (shared weights) conditional generator module $\bbg$, as depicted in Figure~\ref{fig:arc}. 
The architecture of the generator is adapted from Isola et al.~\shortcite{isola2017image} to accept a pose volume as input.
The first, \emph{paired data branch} (blue) of our network is aimed at learning the inverse mapping $\bbp^{-1}$ from the target actor poses to the reference video frames, supervised using paired training data.
The second, \emph{unpaired data branch} (red) utilizes temporal consistency and adversarial losses in order to learn to translate unpaired pose sequences extracted from one or more additional videos into temporally coherent frames that are ideally indistinguishable from those contained in the reference video. 


\paragraph{Paired Data Branch}
The goal of this branch is to train $\bbg$ to approximate the inverse transform, $\bbp^{-1}:Y\rightarrow X$, which maps the paired poses $P(x_i)\in Y$ back to their origin reference video frames $x_i\in X$.
Here we use a reconstruction loss that consists of two parts. The first part is defined as the $\ell_1$ distance between the low-level features extracted from the reconstructed and the ground truth frames by the pretrained VGG-19 network \cite{simonyan2014very}:
\begin{equation}
\Loss_{\text{vgg}}(\bbg) = \bbe_{x\sim\pp
_X}\Vert\phi(\bbg(\bbp(x))) - \phi(x)\Vert_1,
\label{eq:paired_loss}
\end{equation}
where $\phi(\cdot)$ denotes the VGG-extracted features.
In practice, we use the features from the  {\tt relu1\_1} layer, which has the same spatial resolution as the input images.
Minimizing the distance between these low-level VGG features implicitly requires similarity of the underlying native pixel values, but also forces $\bbg$ to better reproduce small scale patterns and textures, compared to a metric defined directly using RGB pixels

\begin{figure}
\centering
\includegraphics[width=1.0\linewidth]{\figures 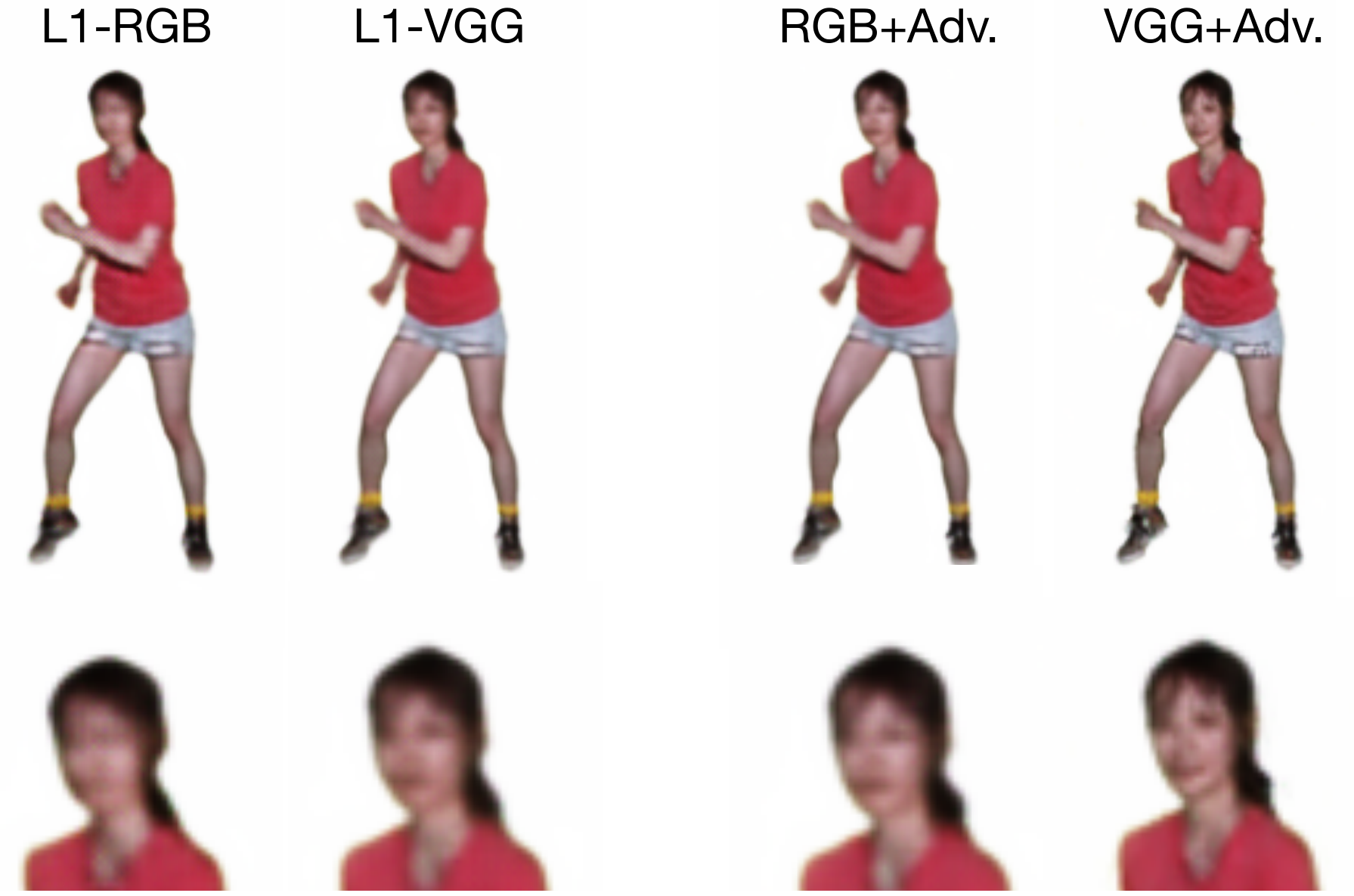}
\caption{Reconstruction loss. Comparison between $\ell_1$ loss on RGB values compared to low-level features extracted by VGG-19. Adding an adversarial loss for paired data \cite{isola2017image} yields the two results on the right. }
\label{fig:vggVsL1}
\end{figure}

For the second part of the loss we employ a Markovian discriminator \cite{isola2017image,li2016precomputed}, $\Dp(y,x)$, trained to tell whether frame $x$ conditioned by pose $y=\bbp(x)$ is real or fake. This further encourages $\bbg$ to generate frames indistinguishable from the target actor in the provided input pose. The discriminator is trained to minimize a standard conditional GAN loss:
\begin{eqnarray}
\Loss_{{\text{gan}}_{\text{p}}}(\Dp) & = & \bbe_{x\sim\pp_X} \left[ \log \Dp(\bbp(x),x)\right] \\ \nonumber
&+& \bbe_{x\sim\pp_X} \left[ \log (1-\Dp(\bbp(x),\bbg(\bbp(x))))\right],
\end{eqnarray}
while the generator is trained to minimize the negation of the second term above.
The two terms together constitute the reconstruction loss of the paired branch with
\begin{equation}
\Loss_{\text{rec}} = \Loss_{\text{gan}_\text{p}} + \lambda_\text{vgg}\Loss_{\text{vgg}}.
\end{equation}
A constant weight of $\lambda_\text{vgg}=10$ is used to balance between the two parts.
Figure~\ref{fig:vggVsL1} shows a comparison between the loss in \eqref{eq:paired_loss} and na{\"i}ve $\ell_1$ loss on the RGB pixel values, while training our system using (only) the paired branch. 
It may be seen that the latter loss results in blurrier images and does not prioritize high frequency details, while $\Loss_{\text{vgg}}$ produces better reconstructions. The advantage of $\Loss_{\text{vgg}}$ is retained after adding the adversarial component of the loss.

\paragraph{Unpaired Data Branch}
The purpose of this branch is to improve the generalization capability of the generator.
The goal is to ensure that the generator is able to translate sequences of unseen poses, extracted from a driving video, into sequences of frames that look like reference video frames, and are temporally coherent.

The training data in this branch is unpaired, meaning that we do not provide a ground truth frame for each pose.
Instead, the input consists of temporally coherent sequences of poses, extracted from one or more driving videos.
The loss used to train this branch is a combination of a temporal coherence loss and a Markovian adversarial loss (PatchGAN) \cite{isola2017image,li2016precomputed}.

The temporal coherence loss is computed using optical flow fields extracted from pairs of successive driving video frames. 
Ideally, we'd like the flow fields between pairs of successive generated frames to be the same, indicating that their temporal behavior follows that of the input driving sequence.
However, we found that comparing smooth and textureless optical flow fields is not sufficiently effective for training. Similarly to Chen et al.~\shortcite{chen2017coherent}, we enforce temporal coherency by using the original optical flow fields to warp the generated frames, and comparing the warped results to the generated consecutive ones.

More formally, let $y_i, y_{i+1}$ denote a pair of consecutive poses, and $f_i$ denote the optical flow field between the consecutive frames that these poses were extracted from.
The temporal coherence loss is defined as:
\begin{equation}
\Loss_{{\text{tc}}}(\bbg) = 
\bbe_{y\sim\pp
	_Y}\left\Vert \alpha(y_{i+1})\cdot\left(\bbw(\bbg(y_i), f_i) -\bbg(y_{i+1})\right) \right\Vert_1,
\label{eq:coherence_loss}
\end{equation}
where $\bbw(x,f)$ is the frame obtained by warping $x$ with a flow field $f$, using bi-linear interpolation. Due to potential differences in body size between the driving and target actors, the differences between the frames are weighted by the spatial map $\alpha(y_{i+1})$, defined by a Gaussian falloff from the limbs connecting the joint locations. 
In other words, the differences are weighted more strongly near the skeleton, where we want to guarantee smooth motion. 

Similarly to the paired branch, here we also apply an adversarial loss. However, here the discriminator, $\Ds(x)$, receives a single input and its goal is to tell whether it is a real frame from the reference video, or a generated one, without being provided the condition $y$:
\begin{equation}
\Loss_{{\text{gan}}_{\text{s}}}(\Ds) = \bbe_{x\sim\pp_X} \left[ \log \Ds(x)\right] +  \bbe_{y\sim\pp_Y}\left[ \log (1-\Ds(\bbg(y)))\right].
\end{equation} 
In addition, another goal of $\Ds$ is to prevent over smoothing effects that might exist due to the temporal loss.
Because the poses come from other videos, they may be very different from the ones in the reference video.
However, we assume that, given a sufficiently rich reference video, the generated frames should consist from pieces (patches) from the target video frames.
Thus, we use a PatchGAN discriminator \cite{isola2017image,li2016precomputed}, which determines whether a frame is real by examining patches, rather than the entire frame. 

\paragraph{Training Specifics}
The two branches constitute our complete network, whose total loss is given by
\begin{equation}
\Loss_{\text{total}} = \Loss_{\text{rec}} + \lambda_{\text{s}} \left( \Loss_{\text{gan}_\text{s}} +\lambda_{\text{tc}} \mathcal{L}_{\text{tc}}\right),
\end{equation}
where $\lambda_\text{s}$ is a global weight for the unpaired branch, and $\lambda_\text{tc}$ weighs the temporal coherency term.

As mentioned earlier, the architecture of the generator $\bbg$ is  adapted from Isola et al.~\shortcite{isola2017image}, by extending it to take pose sequences as input, instead of images. The input tensor is of size $JN\times H\times W$, where $N$ is the number of consecutive poses and the output tensor is $3N\times H\times W$. For all the results in this paper we used $N = 2$ (pairs of poses). The network progressively downsamples the input tensor, encoding it into a compact latent code and then decodes it into a reconstructed frame.
Both of the discriminators, $\Dp$ and $\Ds$ share the same PatchGAN architecture \cite{isola2017image},
but were extended to support space time volumes. The only difference between the two discriminators is that $\Dp$ is extended to receive $N(3+J)$ input channels while $\Ds$ receives $3N$ input channels, since it does not receive a conditional pose with each frame.


In the first few epochs, the training is only done by the paired branch using paired training data (Section~\ref{subsec:data}). In this stage, we set $\lambda_s=0$, letting the network learn the appearance of the target actor, conditioned by a pose. Next, the second branch is enabled and joins the training process using unpaired training data. Here, we set $\lambda_s=0.1$ and $\lambda_{\text{tc}}=10$. 
This stage ensures that the network learns to translate new motions containing unseen poses into smooth and natural video sequences. 

We use the Adam optimizer \cite{kinga2015method} and a batch size of 8.
Half of each batch consists of paired data, while the other half is unpaired. Each half is fed into the corresponding network branch. 
The training data typically consists of about 3000 frames. We train the network for 150 epochs using a learning rate of 0.0002 and first momentum of 0.999. The network is trained from scratch for each individual target actor where the weights are randomly initialized based on a Normal distribution $\mathcal{N}(0, 0.2)$.

Typically, 3000 frames of a target actor video, are sufficient to train our network. However, 
the amount of data is only one of the factors that one should consider. A major factor that influences the results is our parametric model. Unlike face models \cite{Thies2016face2face} which explicitly control many aspects, such as expression and eye gaze, our model only specifies the positions of joints. Many details, such as hands, shoes, face, etc., are not explicitly controlled by the pose. Thus, given a new unseen pose, the network must generate these details based on the connections between pose and appearance that has been seen in the paired training data. Hence, unseen input poses that are similar to those that exist in the paired data set, will yield better performance. In the next section, we propose a new metric suitable for measuring the similarity between poses in our context.


%

\section{Pose Metrics}
\label{sec:metric}


As discussed earlier, the visual quality of the frames generated by our network, depends strongly on the similarity of the poses in the driving video to the poses present in the paired training data, extracted from the reference video.
However, there is no need for each driving pose to closely match a single pose in the training data. Due to the use of a PatchGAN discriminator, a frame generated by our network may be constructed by combining together local patches seen during the training process. Thus, similarity between driving and reference poses should be measured in a local and translation invariant manner.
In this section we propose a pose metrics that attempt to measure similarity in such a manner.



\begin{figure}
\centering
\includegraphics[width=1.0\linewidth]{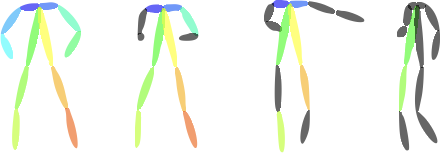}
\begin{tabular}{cccc}
\hspace{0.1cm} (a) $d=0$ &  \hspace{0.6cm}(b) $d=4.2$ &  \hspace{0.6cm}(c) $d=8.5$ &  \hspace{0.6cm}(d) $d=17.2$
\end{tabular}
\caption{Pose to pose metric: (a) is a reference pose, while (c)--(d) are different poses from the driving video. The differences between the poses, as well as the values of our metric increase from left to right. Limbs whose distance to their counterparts in (a) exceeds the threshold of 8 are shown in gray.
Thus, our metric helps detect limbs that are likely to be hard to reconstruct.}
\label{fig:skeletons}
\end{figure}

\paragraph{Skeleton Descriptor}
As explained in Section~\ref{subsec:data}, poses are extracted as stacks of part confidence maps \cite{cao2017realtime}. A 2D skeleton may be defined by estimating the position of each joint (the maximal value in each confidence map), and connecting the joints together into a predefined set of limbs.
We consider 12 limbs (3 for each leg and 3 for each arm and shoulder), as depicted in Figure~\ref{fig:skeletons}, for example.
To account for translation invariance, our descriptor consists of only the orientation and length of each limb.  The length of the limb is used to detect difference in scale that might affect the quality of the results, and the orientation is used to detect rotated limbs that haven't been seen by the network during training. Fang et al.~\shortcite{fang2018weakly} measure the distance between two poses using the normalized coordinates of the joints. This means that two limbs of the same size and orientation but shifted, will be considered different. In contrast, our descriptor is invariant to translation.  

Specifically, our skeleton descriptor consists of vectors that measure the 2D displacement (in the image plane) between the beginning and the end of each limb. Namely, for every limb $l\in L$ we calculate $\Delta x^l = x^{l}_{1} - x^{l}_{2}$ and $\Delta y^l = y^{l}_{1} - y^{l}_{2}$, where $(x^{l}_{1},y^{l}_{1})$ and $(x^{l}_{2},y^{l}_{2})$ are the image coordinates of the first and second joint of the limb, respectively.  
The descriptor $p_i$ consist of $L = 12$ such displacements: $p_i = \{p^1_i, \ldots, p^L_i \}$, where $p^l_i = (\Delta x_i^l, \Delta y_i^l)$.

\paragraph{Pose to Pose Metric}
Given two pose skeleton descriptors, $p_i,p_j$, we define the distance between their poses as the average of distances between their corresponding limbs,
\begin{equation}
\label{eq:p2p}
	d(p_i,p_j) = \frac{1}{|L|}\sum_{l\in L} \left\|p_i^l - p_j^l \right\|_2.
\end{equation}

Figure~\ref{fig:skeletons} depicts the above distance between a reference pose (a) and different driving poses (b)--(d). Note that as the driving poses diverge from the reference pose, the distances defined by Eq.~\eqref{eq:p2p} increase. Limbs whose distance is above a threshold, $d_l(p_i,p_j) > \gamma$, are indicated in gray. These limbs are unlikely to be reconstructed well from the frame corresponding to the pose in (a). Other limbs, which have roughly the same orientation and length as those in (a) are not indicated as problematic, even when significant translation exists (e.g., the bottom left limb of pose (c)).

\paragraph{Pose to Sequence Metric}
Using the above metric we can estimate how well a given driving pose $p_i$, might be reconstructed by a network trained using a specific reference sequence $v$. The idea is simply to find, for each limb in $p_i$, its nearest neighbor in the entire sequence $v$, and compute the average distance to these nearest-neighbor limbs:
\begin{equation}
\label{eq:p2v}
d(p_i,v) = \frac{1}{|L|}\sum_{l\in L}\min_{p_j\in v}\left\|p_i^l - p_j^l \right\|_2.
\end{equation}
In other words, this metric simply measures whether each limb is present in the training data with a similar length and orientation. It should be noted that the frequency of the appearance of the limb in the training data set is not considered by this metric. Ideally, it should be also taken into account, as it might influence on the reconstruction quality.

\begin{figure}
\centering
\includegraphics[width=1.0\linewidth]{\figures 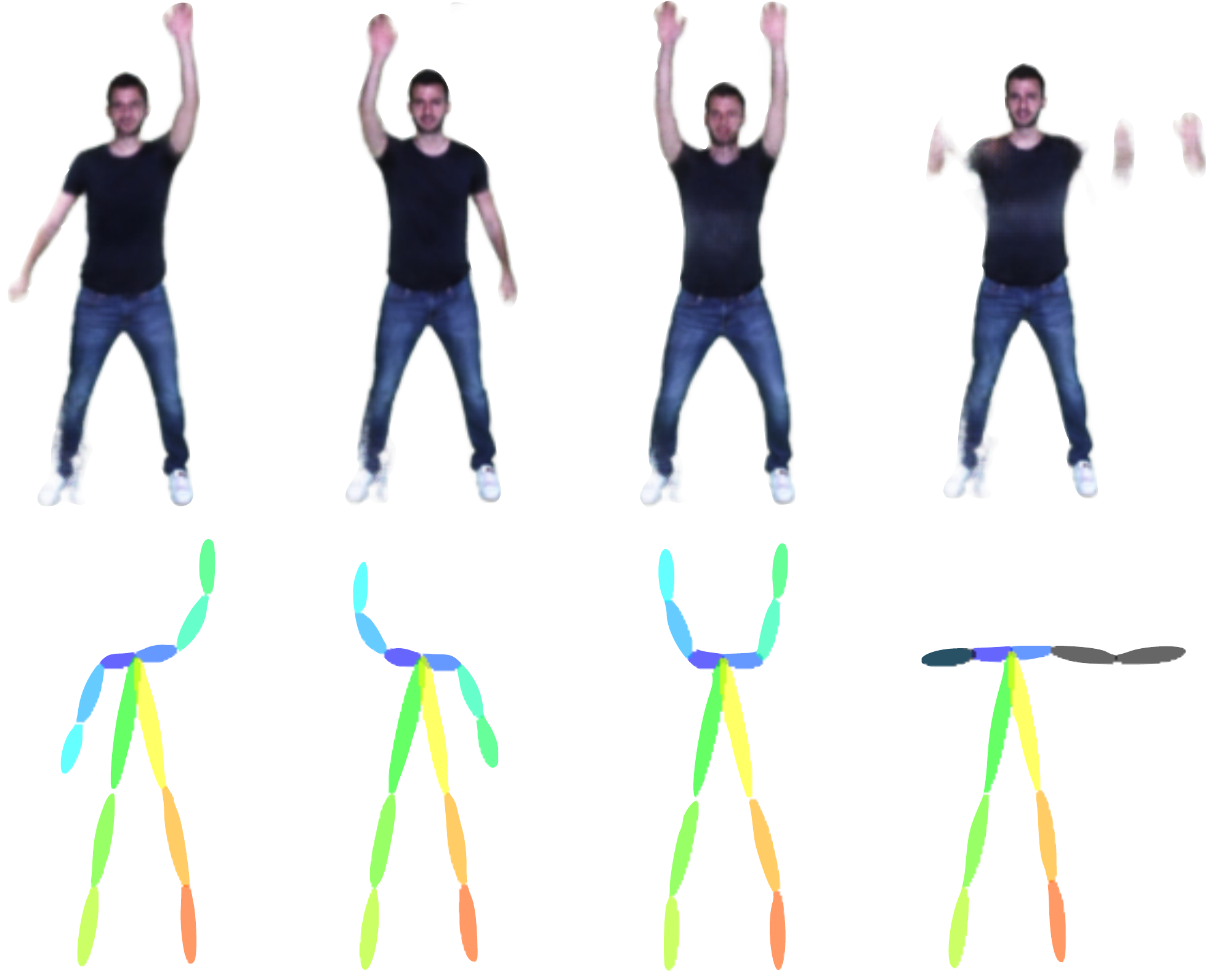}
\begin{tabular}{cccc}
(a) 0.1 &  \hspace{0.7cm} (b) 0.1 &  \hspace{0.7cm} (c) 0.1 &  \hspace{0.7cm}(d) 5.2
\end{tabular}
\caption{Pose to sequence metric. Here, the paired training data only features vertical arm motion, with a single arm being raised at a time. Our metric yields small values for poses where only one arm is raised (a,b), as well as for one with both arms raised (c). Even though such a pose has not been seen during training, our network is able to reconstruct it. However, we fail to reconstruct a pose where the arms are stretched horizontally (d), and indeed the value of our metric is high in this case.}
\label{fig:hands}
\end{figure}

The above metric is demonstrated in Figure \ref{fig:hands}, (a)--(c) show well reconstructed frames from poses whose distance to the reference sequence is small. The pose in (d) does not have good nearest neighbors for the arm limbs, since the reference video did not contain horizontal arm motions. Thus, the reconstruction quality is poor, as predicted by the larger value of our metric.

However, it should be emphasized that small pose distances alone cannot guarantee good reconstruction quality. Additional factors should be considered, such as differences in body shape and clothing, which are not explicitly represented by poses, as well as the accuracy of pose extraction, differences in viewpoint, and more.

\section{Results and Evaluation}
\label{sec:results}

We implemented our performance cloning system in PyTorch, and performed a variety of experiments on a PC equipped with an Intel Core i7-6950X/3.0GHz CPU (16 GB RAM), and an NVIDIA GeForce GTX Titan Xp GPU (12 GB).
Training our network typically takes about 4 hours for a reference video with a 256$\times$256 resolution.
When cloning a performance, the driving video frames must be first fed forward through the pose extractor $\bbp$ at 35\emph{ms} per frame, and the resulting pose must be fed forward through the generator $\bbg$ at 45\emph{ms} per frame. Thus, it takes a total of 80\emph{ms} per cloned frame, which means that a 25fps video performance may be cloned in real time if two or more GPUs are used.

To carry out our experiments, we captured videos of 8 subjects, none of which are professional dancers. The subjects were asked first to imitate a set of simple motions (stretches, boxing motions, walking, waving arms) for two minutes, and then perform a free-style dance for one more minute in front of the camera.
The two types of motions are demonstrated in the supplemental video. In all these videos we extracted the performer using an interactive video segmentation tool \cite{fan2015jumpcut}, since this is what our network aims to learn and to generate. We also gathered several driving videos from the Internet, and extracted the foreground performance from these videos as well. 

\begin{figure}
	\centering
	\includegraphics[width=1.0\linewidth]{\figures 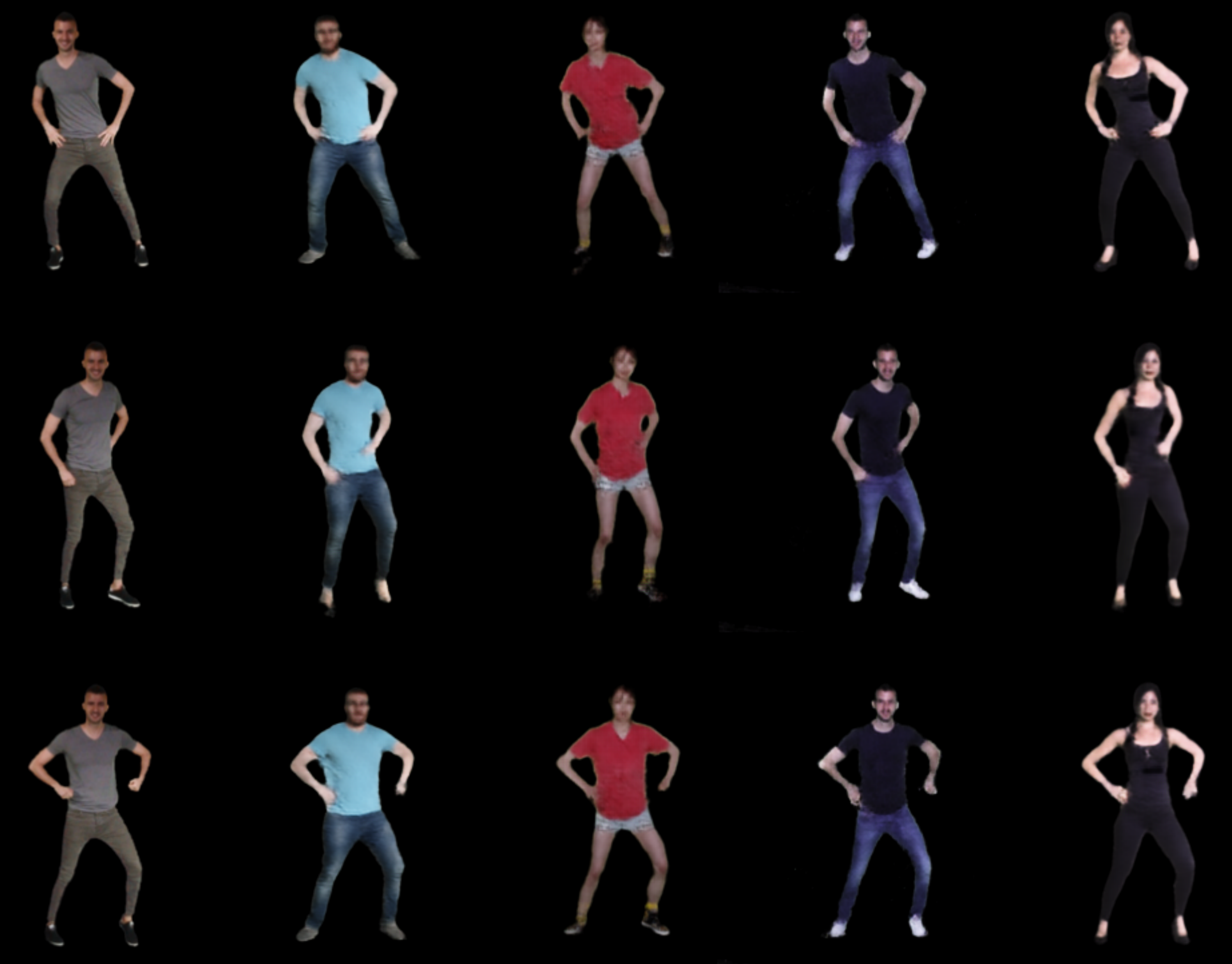}
	\caption{Leftmost column: three driving frames from which poses were extracted. The other four columns show 4 different target actors cloning these three poses. }
	\label{fig:gallery}
\end{figure}

Figure \ref{fig:gallery} shows four of our subjects serving as target individuals for cloning three driving poses extracted from the three frames in the leftmost column. The supplementary video shows these individuals cloning the same dance sequence with perfect synchronization.

Below, we begin with a qualitative comparison of our approach to two baselines (Section \ref{sec:qualitative}), and then proceed to a quantitative evaluation and comparison with the baselines (Section \ref{sec:quantitative}).
It should be noted that, to our knowledge, there are no existing approaches that directly target video performance cloning. Thus, the baselines we compare to consist of existing state-of-the-art methods for unpaired and paired image translation.

\subsection{Comparison to Other Algorithms}
\label{sec:qualitative}

We compare our video performance cloning system with state-of-the-art methods for unpaired and paired image translation techniques. The comparison is done against image translation methods since there is no other method that we're aware of that is directly applicable to video performance cloning. 
Thus, we first compare our results to CycleGAN \cite{zhu2017unpaired}, which is capable of learning a mapping between unpaired image sets from two domains. In our context, the two domains are frames depicting the performing actor from the driving video and frames depicting the target actor from the reference video.

\begin{figure}
	\centering
	\includegraphics[width=1.0\linewidth]{\figures 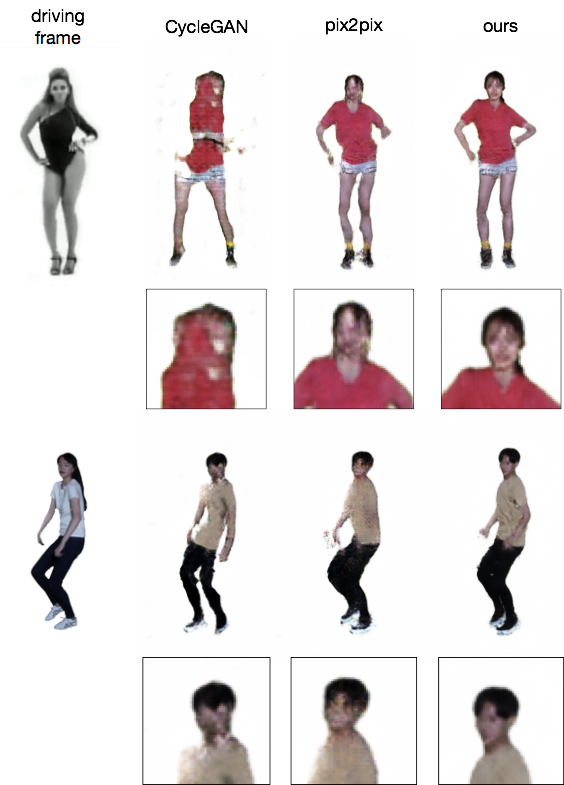}
	\caption{Qualitative comparison between CycleGAN \cite{zhu2017unpaired}, pix2pix \cite{isola2017image}, and our method (rightmost column). }
	\label{fig:comparison}
\end{figure}

It is well known that CycleGAN tends to transform colors and texture of objects, rather than modifying their geometry, and thus it tends to generate geometrically distorted results in our setting. More specifically, it attempts to replace the textures of the driving actor by those of the target one, but fails to properly modify the body geometry. This is demonstrated in Figure~\ref{fig:comparison} (2nd column from the left).

Next, we compare our results to the pix2pix framework \cite{isola2017image}. This framework learns to translate images using a paired training dataset, much like we do in our paired branch. To apply it in our context, we extend the pix2pix architecture to support translation of 3D part confidence maps, rather than images, and train it using a paired set of poses and frames from the reference video.
Since the training uses a specific and limited set of motions, it may be seen that unseen poses cause artifacts. In addition, temporal coherency is not taken into consideration and some flickering exist in the results, which are included in the supplemental video.

Figure~\ref{fig:comparison} shows a qualitative comparison between the results of the two baselines described above and the results of our method (rightmost column). It may be seen that our method generates frames of higher visual quality. The driving frame in this comparison is taken from a music video clip found on the web.

\begin{figure}
	\centering
	\includegraphics[width=1.0\linewidth]{\figures 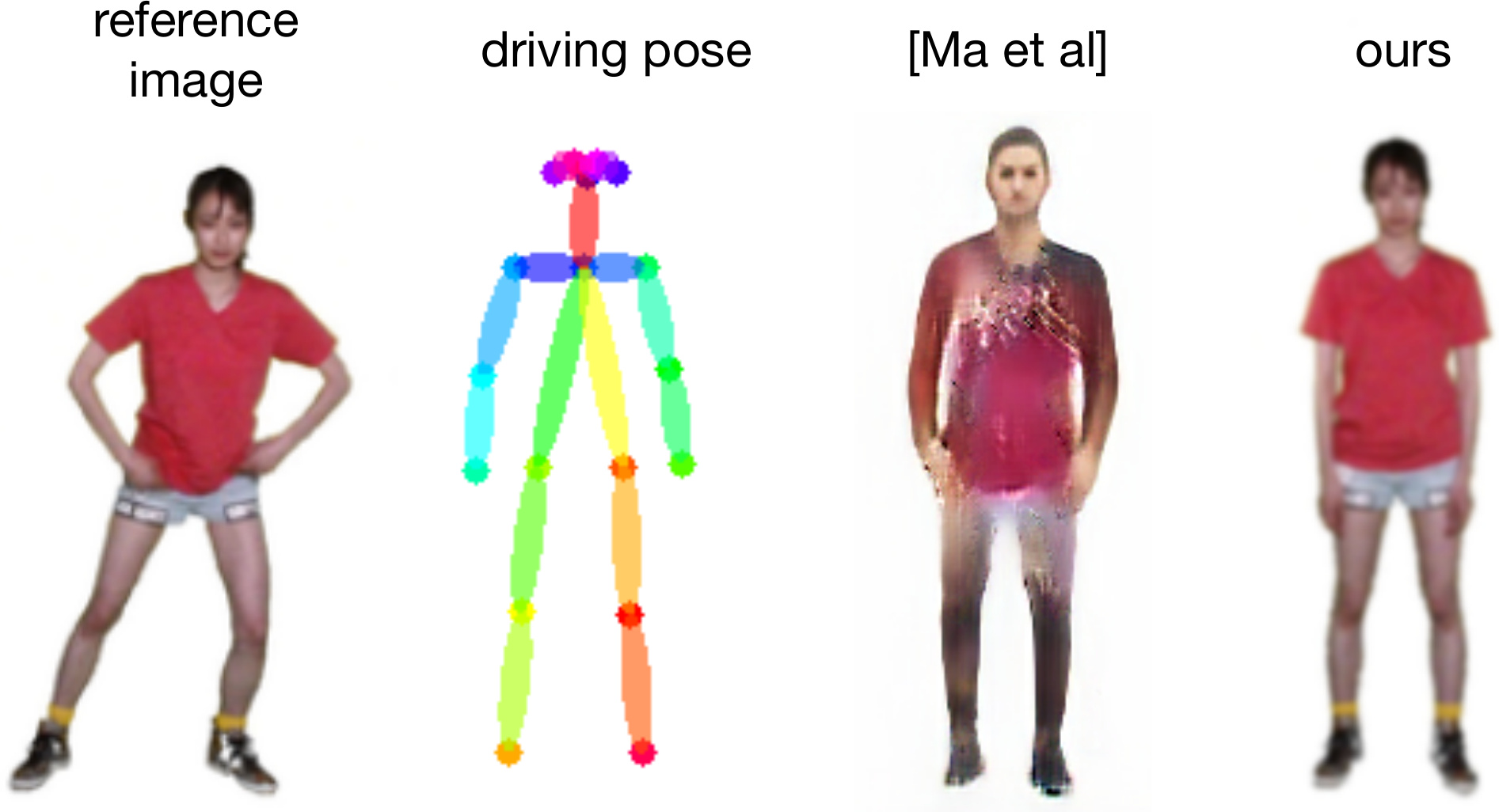}
	\caption{Qualitative comparison with pose-guided person image generation \cite{ma2017pose}.}
	\label{fig:ma}
\end{figure}

We next compare our method to Ma el al. \shortcite{ma2017pose}, which is aimed at person image generation conditioned on a reference image and a target pose. We used a pertained version of their network that was trained on the ``deep fashion'' dataset. This dataset consists of models in different poses on white background, which closely resembles our video frames. Figure~\ref{fig:ma} compares a sample result of
Ma et al. to our result for the same input pose using our network, trained on a reference video of the depicted target actor. It may be seen that the model of Ma et al. does not succeed in preserving the identity of the actor. The result of their model will likely improve after training with the same reference video as our network, but it is unlikely to be able to translate a pose sequence into a smooth video, since it does not address temporal coherence.

\subsection{Quantitative Evaluation}
\label{sec:quantitative}

We proceed with a quantitative evaluation of our performance cloning quality. 
In order to be able compare our results to ground truth frames, we evaluate our approach in a self-reenactment scenario. Specifically, given a captured dance sequence by the target actor, we use only two thirds of the video to train our network. The remaining third is then used as the driving video, as well as the ground truth result.
Ideally, the generator should be able to reconstruct frames that are as close as possible to the input ones.

\begin{figure}
	\centering
	\includegraphics[width=1.0\linewidth]{\figures 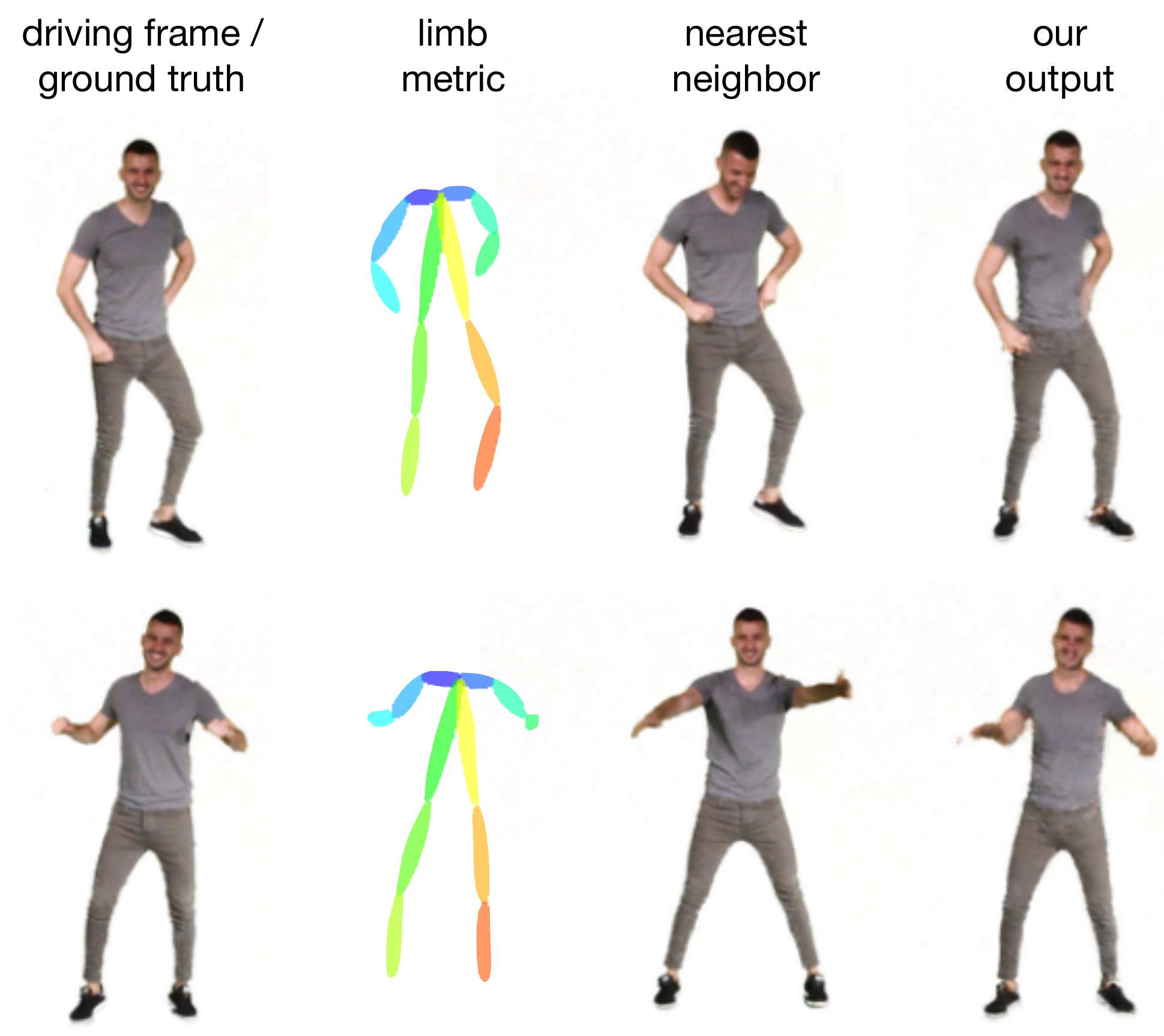}
	\caption{Self-reenactment test. Our translation network is trained on the first two thirds of a video sequence of a given target actor and tested  on the remaining last third. The frames produced by our method are visually closer to the input/ground truth than any of the training data.}
	\label{fig:reenact}
\end{figure}

We have observed that although the last third of the video captures the same individual dancing as the first two thirds, which were used for training, the target actors seldom repeat the exact same poses. Thus, the generator cannot merely memorize all the frames seen during training and retrieve them at test time. This is demonstrated qualitatively in Figure~\ref{fig:reenact}: it may be seen that for the two shown driving poses, the nearest neighbors in the training part of the video are close, but not identical. In contrast, the frames generated by our system reproduce the driving pose more faithfully.

We performed the self-reenactment test for three different actors, and report the mean square error (MSE) between the RGB pixel values of the ground truth and the generated frames in Table~\ref{tab:RGBerror}. We also report the MSE for frames generated by the two baselines, CycleGAN \cite{zhu2017unpaired} and pix2pix \cite{isola2017image}. Our approach consistently results in a lower error than these baselines. As a reference, we also report the error when the self-reenactment is done using the nearest-neighbor poses extracted from the training data. As pointed out earlier, the target actors seldom repeat their poses, and the errors are large.


\begin{table}
	\caption{The mean square error (MSE) of RGB values between the fake output of our network and and the ground truth in a reenactment setting of three different actors. We compare our method to CycleGAN \cite{zhu2017unpaired} and pix2pix \cite{isola2017image}.}
	\begin{tabular}{c|c|c|c|c}
		&Nearest Neighbor &CycleGAN & pix2pix & Ours\\
		\hline
		\hline
		actor1 & 26.4 & 28.4 & 20.8 & \bf{15.4}\\
		actor2 & 29.4 & 27.2 & 24.8 & \bf{16.9}\\
		actor3 & 28.4 & 32.9 & 19.9 & \bf{16.3}\\
	\end{tabular}
	\label{tab:RGBerror}
\end{table}

\begin{figure}
	\centering
	\includegraphics[width=0.75\linewidth]{\figures 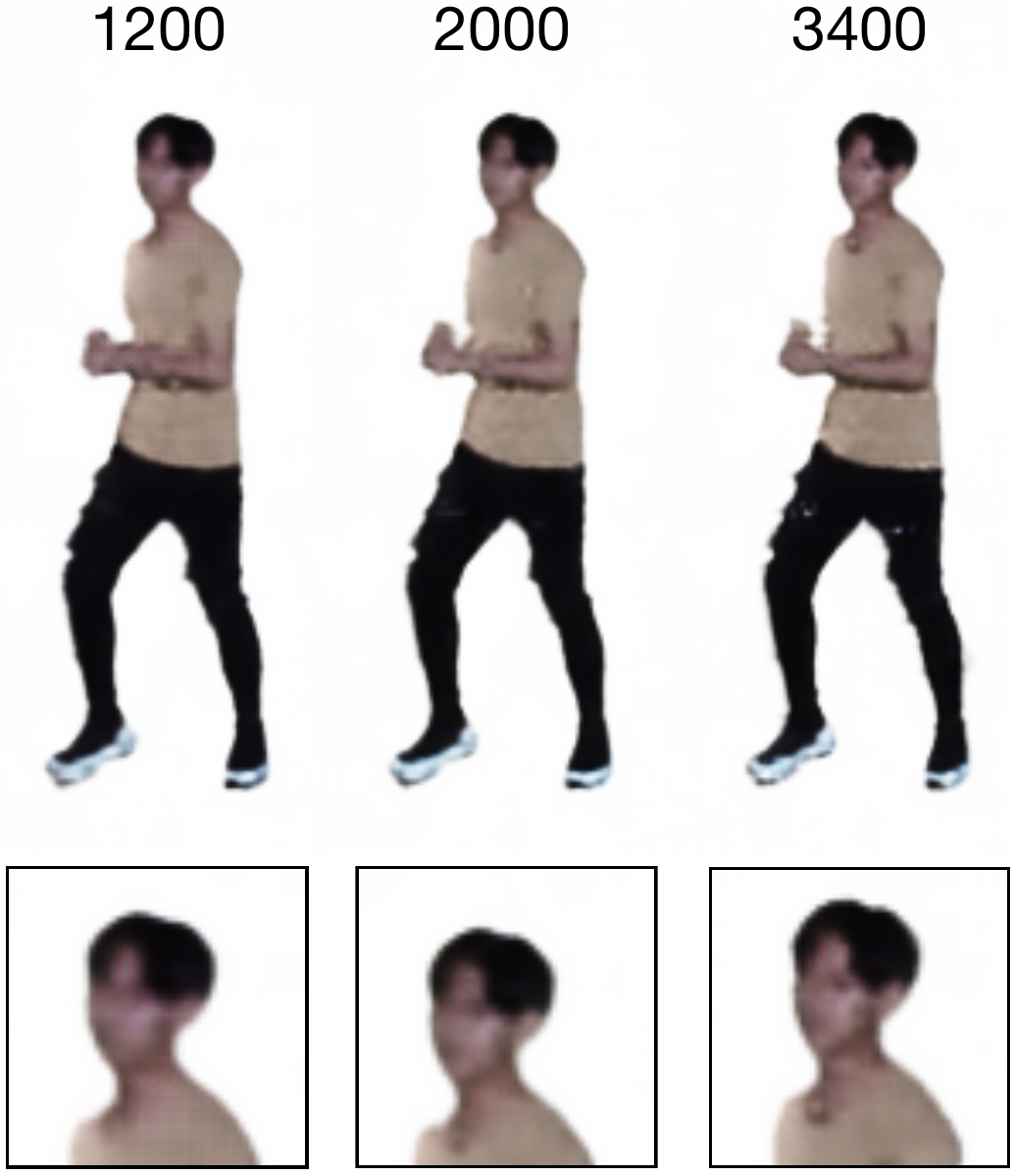}
	\caption{Impact of the training set size (on the self-reenactment test). The visual quality of the results increases with the number of training pairs.}
	\label{fig:set-size}
\end{figure}

We also examine the importance of the training set. In this experiment, we trained our translation network with 1200, 2000, and 3400 frames of the reference sequence of one of the actors (actor1). 
The visual quality of the results is demonstrated in Figure~\ref{fig:set-size}. 
The MSE error is 18.5 (1200 frames), 16.7 (2000 frames), and 15.4 (3400 frames). Not surprisingly, the best
results are obtained with the full training set.

Finally, we examine the benefit of having the unpaired branch. We first train our network in a self reenactment setting, as previously described, using only the paired branch. Next, we train the full configuration, and measure the MSE error for both versions. We repeat the experiment for 3 different reference videos and the reported errors, on average, are 17.5 for paired branch training only vs. 16.2 when both branches are used. Visually, the result generated after training with only the paired branch exhibit more temporal artifacts. These artifacts are demonstrated in the supplementary video.

\subsection{Online performance cloning}
\label{sec:online}

Once our network has been properly trained, we can drive the target actor by feeding the network with unseen driving videos. In particular, given the quick feed-forward time, with two or more GPUs working in parallel it should be possible to clone a driving performance as it is being captured, with just a small buffering delay. Naturally, the quality of the result depends greatly on the diversity of the paired data that the network was trained with.

\section{Discussion and Future work}
\label{sec:discussion}
We have presented a technique for video-based performance cloning. We achieve performance cloning by training a generator to produce frame sequences of a target performer that imitate the motions of an unseen driving video. The key challenge in our work is to learn to synthesize video frames of the target actor that was captured performing motions which may be quite different from those performed in the driving video. Our method leverages the recent success of deep neural networks in developing generative models for visual media. However, such models are still in their infancy, and they are just beginning to reach the desired visual quality. This is manifested by various artifacts when generating fine details, such as in human faces, or in regions close to edges or high frequencies. Recent efforts in progressive GANs provide grounds for the belief that stronger neural networks, trained on more and higher resolution data will eventually be able to produce production quality images and videos. 

Our method mainly focuses on two challenging aspects, dealing with unseen poses and generating smooth motions. Nevertheless, there is of course a direct correlation between the synthesis success and the similarity between the poses in the reference video and the driving videos. Moreover, the sizes and proportions of the performers' skeletons should also be similar. With our current approach, it is impossible to transfer the performance of a grownup to a child, or vice versa.

Our current implementation bypasses several challenging issues in dealing with videos. We currently do not attempt to generate the background, and extract the performances from the reference video using an interactive tool. Furthermore, we heavy rely on the accuracy of the pose extraction unit, which is a non-trivial system on its own, and is also based on a deep network, that has its own limitations. 

Encouraged by our results, we plan to continue and improve various aspects, including the ones mentioned above. In particular, we would like to consider developing an auxiliary universal adversarial discriminator that can tell how likely a synthesized frame is or how realistic a synthesized motion is. Such a universal discriminator might be trained on huge human motion repositories like the CMU motion capture library. We are also considering applying local discriminators that focus on critical parts, such as faces, hands, and feet. Synthesizing the fine facial details is important for the personal user experience, since we believe that the primary application of our method is a tool that enhances an amateur performance to match a perfect one, performed by a professional idol, as envisioned in beginning of this paper.

\section{acknowledgements}
\label{sec:acknowledgements}
We would like to thank the various subjects (Oren, Danielle, Adam, Kristen, Yixin, Fang) that were part of the data capturing and helped us to promote our research by cloning their performances.

\bibliographystyle{ACM-Reference-Format}
\bibliography{references} 

\end{document}